\documentclass[runningheads]{llncs}

\usepackage{eccvabbrv}

\usepackage{graphicx}
\usepackage{booktabs}
\usepackage{multirow}
\usepackage[ruled,vlined]{algorithm2e}
\usepackage[accsupp]{axessibility}  

\usepackage{orcidlink}
\usepackage{booktabs}
\usepackage{multirow}
\usepackage{array}
\usepackage{booktabs}
\usepackage{pifont}

\begin{document}

\title{Online Reasoning Video Object Segmentation} 

\titlerunning{Online Reasoning Video Object Segmentation}

\author{Jinyuan Liu\inst{1,2} \and
Yang Wang\inst{1} \and
Zeyu Zhao\inst{3} \and
Weixin Li\inst{2} \and
Song Wang\inst{1} \and
Ruize Han\inst{1}}

\authorrunning{Liu et al.}

\institute{Faculty of Computer Science and Artificial Intelligence, Shenzhen University of Advanced \and
School of Computer Science and Engineering, Beihang University \and School of Electrical and Information Engineering, Northeast Petroleum University
}
\maketitle

\begin{abstract}
Reasoning video object segmentation predicts pixel-level masks in videos from natural-language queries that may involve implicit and temporally grounded references. However, existing methods are developed and evaluated in an offline regime, where the entire video is available at inference time and future frames can be exploited for retrospective disambiguation, deviating from real-world deployments that require strictly causal, frame-by-frame decisions. We study Online Reasoning Video Object Segmentation (ORVOS), where models must incrementally interpret queries using only past and current frames without revisiting previous predictions, while handling referent shifts as events unfold. To support evaluation, we introduce ORVOSB, a benchmark with frame-level causal annotations and referent-shift labels, comprising 210 videos, 12,907 annotated frames, and 512 queries across five reasoning categories. We further propose a baseline with continually-updated segmentation prompts and a structured temporal token reservoir for long-horizon reasoning under bounded computation. Experiments show that existing methods struggle under strict causality and referent shifts, while our baseline establishes a strong foundation for future research.
\end{abstract}

\section{Introduction}
\label{sec:intro}

\textit{A video is not a sequence of images -- it is a temporal narrative. Remove the flow, and you erase its meaning.} {\par {\raggedleft {---Andr\'{e} Bazin} \par}}

Reasoning video object segmentation~\cite{yan2024visa,gong2025devil,munasinghe2025videoglamm} aims to temporally predict pixel-level masks in videos from natural-language queries, where the referential relationship is implicit requiring logical reasoning.
With the rapid progress of multimodal large language models (MLLMs)~\cite{liu2023visual,jin2024chat,bai2023qwen}, recent work~\cite{yan2024visa} has established a unified paradigm that maps video frames and language prompts to segmentation tokens for mask decoding, improving the ability of open-world grounding and compositional reasoning.

Despite this progress, existing methods~\cite{pont20172017,seo2020urvos,ding2023mevis,yan2024visa} are all developed and evaluated in an offline regime, where the entire video is available at inference stage.
This setting has two significant limitations.
First, \textbf{retrospective disambiguation}: the model may use future observations to identify the referent and then interpret earlier frames with hindsight~\cite{niu2025ovo}. For instance, a vehicle may begin to move at the middle moment of the video, it will be segmented along the whole video, including the first half of the video when it remains stationary.
Second, \textbf{online application limitation}: while suitable for post-hoc analysis, existing setting does not explicitly assess the frame-by-frame, causal decision making required in real deployments such as robotics~\cite{obrenovic2025generative}, embodied agents~\cite{yang2025embodiedbench}, and on-device video analytics~\cite{dai2024context}, where frames arrive sequentially and future information is unavailable~\cite{bothra2023veritas}.

From the above observations, in this work, we propose to study a new and practical problem, i.e., online reasoning video object segmentation (\textbf{ORVOS}).
Specifically, the model must interpret the query incrementally using only past and current frames, without revisiting or revising previous predictions~\cite{huang2025online}.
Although practical, the online setting also brings new challenges for the reasoning video object segmentation:
i) \textbf{Strict Causality}: the referred target can be determinate and segmented only after relevant actions or interactions occur.
For instance, `the moving vehicle' can not be segmented until it starts moving.
ii) \textbf{Referent Shifts}: many event-driven expressions induce referent shifts, where the queried entity changes over time.
For instance, queries such as `the moving vehicle' or `the animal that is attacked' may correspond to different instances at different stages of an unfolding event.
Such properties introduce challenges beyond global reasoning and temporal association in classical RVOS, including causal disambiguation, progressive interpretation, and shift-aware target tracking.


Considering the above problem characteristics and challenges, existing works mainly encounter the following problems.
On the one hand, existing benchmarks are not designed to evaluate these online inference capabilities.
Most of the referring or reasoning video object segmentation datasets~\cite{yan2024visa,gong2025devil,munasinghe2025videoglamm,bai2024one,zheng2025villa} assume full-video access and typically treat the referent as fixed for each query.
Consequently, they cannot faithfully measure incremental decision making, adaptation to evolving targets, or robustness under strictly causal inference.
On the other hand, from the modeling perspective, existing MLLM-based approaches are also mostly optimized for offline processing and are not explicitly tailored to incremental inference.
Many pipelines summarize the video into a single representation and reuse it for segmentation: some perform keyframe-level reasoning and then propagate predictions via tracking~\cite{yan2024visa,gong2025devil}, while others segment all frames with a shared video-level feature~\cite{bai2024one}.
More recent methods~\cite{lin2025glus,zheng2025villa} incorporate global--local fusion to obtain frame-level tokens, but typically do not maintain a causal mechanism to accumulate and reuse past disambiguation cues as temporal memory.
This motivates an online design that continually updates the segmentation prompt over time and explicitly stores historical prompt features for progressive disambiguation.

In this work, to establish the study of ORVOS, we build a new benchmark and propose a new baseline method.
Specifically, to \textbf{fill the benchmark gap}, we introduce ORVOSB, a new Benchmark for Online Reasoning Video Object Segmentation.
In ORVOSB, we annotate the referred targets under the strict causality constraint by accurately determining whether the target satisfies the referring queries frame-by-frame. 
Moreover, ORVOSB includes various referent shifts, which require models to resolve ambiguity over time and dynamically adapt as evidence accumulates.
In total, it contains 210 videos and 12,907 annotated frames, together with 512 reasoning queries spanning five categories: attribute, spatial relation, action or state change, interaction, and external knowledge.
To \textbf{build the methodology basics}, we further propose a simple and effective baseline for ORVOS.
The proposed method maintains a continually-updating segmentation prompt: at each time step, the MLLM generates a frame-specific target-aware token conditioned on the current frame, a short window of context frames, and aggregated historical memory, which are together used to construct a history-aware prompt for mask prediction.
To support long-term reasoning under bounded computation, we further maintain a structured temporal token reservoir with dense-to-sparse retention and query-based aggregation.
Experimental results show that state-of-the-art RVOS methods degrade substantially under strict causality and referent shifts on ORVOSB, while the proposed method establishes a strong baseline on ORVOS, which provides a foundation for future research on the proposed causally-grounded and referent-alterable reasoning segmentation task.
In summary, we make the following contributions:
\begin{itemize}
    \item We propose a new and practical problem of online reasoning video object segmentation, which gets rid of the retrospective disambiguation in classical RVOS and can be applied in real-world online applications.
    \item We build ORVOSB, a new benchmark with various referent-shift annotations and causality evaluation protocol, which enables accurate assessment of online causality and target switching.
    \item We propose a simple and effective baseline with past information enhancing and dynamic embedding updating, which uses a continually-updating segmentation prompt and a structured temporal token reservoir for online reasoning mask prediction.
    \item We provide comprehensive experiments that reveal the limitations of existing methods under online RVOS setting and validate the usefulness of the proposed benchmark and the effectiveness of the proposed method.
\end{itemize}

\section{Related Work}
\label{sec:relatedwork}

\subsection{Referring Video Object Segmentation}

Referring video object segmentation (RVOS)~\cite{botach2022end,wu2022language,yan2024referred,ding2023mevis} aims to segment target objects in videos according to explicit natural language expressions that directly describe the target. 
Early approaches primarily rely on multimodal fusion, aligning visual representations with linguistic cues to associate targets across space and time. 
For instance, URVOS~\cite{seo2020urvos} employs cross-modal attention and memory mechanisms to jointly address RVOS and semi-supervised video object segmentation, while Hui \etal~\cite{hui2021collaborative} dynamically recombine linguistic and visual features to highlight spatiotemporal regions of interest.
Inspired by DETR-style architectures~\cite{zhu2020deformable}, query-based end-to-end methods such as MTTR~\cite{botach2022end} and ReferFormer~\cite{wu2022language} leverage language queries to directly guide target localization and mask decoding within a unified framework. 
More recently, MLLM-based approaches~\cite{bai2024one,munasinghe2025videoglamm,yan2024visa,zheng2025villa} further enhance compositional reasoning capability to better handle complex expressions.

\subsection{Reasoning Video Object Segmentation}

Reasoning object segmentation~\cite{lai2024lisa,yang2023lisa++,ren2024pixellm} focuses on implicit and compositional expressions that require higher-level inference, where accurate mask prediction depends on intent understanding and sometimes external knowledge. 
Recent works establish a unified paradigm that connects an MLLM~\cite{liu2023visual,chen2024internvl,jin2024chat} with a mask decoder~\cite{kirillov2023segment,ravi2024sam}, enabling video frames and language queries to be translated into a special \texttt{<SEG>} token that is subsequently decoded into pixel-level masks.
Extending this formulation to videos, VISA~\cite{yan2024visa} performs keyframe-level reasoning using a pretrained MLLM and propagates masks temporally to remaining frames. 
Subsequent works improve spatiotemporal modeling from different perspectives. 
VideoLISA~\cite{munasinghe2025videoglamm} adopts sparse–dense frame sampling to balance temporal coverage and spatial precision. 
VRS-HQ~\cite{gong2025devil} introduces hierarchical tokens for enhanced temporal reasoning, followed by mask propagation. 
GLUS~\cite{lin2025glus} unifies global referring reasoning and local temporal continuity within a transformer framework. 
Despite stronger reasoning capability, these methods typically rely on offline video access.

\subsection{Online Video Understanding}

Recent advances in MLLMs~\cite{alayrac2022flamingo,bai2025qwen3,liu2023visual} extend video understanding from offline processing to real-time streaming scenarios, where frames arrive sequentially under causal constraints. 
VideoLLM-Online~\cite{chen2024videollm} introduces a streaming framework for temporally aligned, long-context video conversation. 
StreamingVLM~\cite{xu2025streamingvlm} aligns training and inference for stable real-time understanding of unbounded visual streams. 
LiveVLM~\cite{ning2025livevlm} reduces inference cost through key-value cache pruning and frame-wise token merging, while StreamChat~\cite{xiong2025streaming} incorporates hierarchical memory for multi-round dialogue. 
VideoStreaming~\cite{qian2024streaming} further enables arbitrary-length video processing with a constant number of adaptively selected tokens.
Although these approaches enable streaming inference, they mainly operate at the language level and focus on conversational understanding. 
In contrast, pixel-level tasks such as video object segmentation require strict spatial-temporal consistency under causal constraints, yet remain underexplored in online settings.

\section{Online Reasoning Video Object Segmentation Benchmark}

\subsection{Motivation}
Most existing referring and reasoning video object segmentation benchmarks~\cite{pont20172017,seo2020urvos,ding2023mevis,yan2024visa} are built around an offline protocol, where the full video is accessible during inference.
As a result, evaluation may implicitly allow retrospective disambiguation with future frames~\cite{niu2025ovo}, rather than measuring strictly causal, frame-by-frame segmentation in online video scenarios~\cite{bothra2023veritas}.

More importantly for dataset design, existing benchmarks~\cite{yan2024visa,zheng2025villa,bai2024one} typically assume a fixed referent for each query.
Masks are annotated with a single target identity across the entire sequence, leaving no supervision for event-driven cases where the semantically correct referent changes over time.
Without explicit referent-shift annotations and their temporal boundaries, current datasets cannot evaluate whether a model adapts its target online, nor can they disentangle failures caused by early ambiguity from those caused by evidence revealed later.

Recent efforts on online video understanding emphasize sequential input and causal constraints~\cite{lin2024streamingbench,xiong2025streaming,yang2025svbench,wang2025streambridge}, but they mainly focus on language-level tasks and do not provide pixel-level annotations or protocols for causal mask prediction.
Therefore, a dedicated benchmark is still missing that simultaneously offers (i) a strictly causal evaluation protocol and (ii) referent-shift annotations to assess online target disambiguation and temporal consistency.

These gaps motivate ORVOSB, a benchmark for online reasoning video object segmentation under strict causality, featuring referent-shift annotations and a causal evaluation protocol.

\subsection{Problem Definition}

We formally define the task of online reasoning video object segmentation (ORVOS). Given a streaming video and a textual query, the model is required to predict the segmentation mask of the referred object at each time step, without accessing any future frame.

\paragraph{Streaming Video Input.}
Let $\mathcal{V} = \{ I_1, I_2, \dots, I_T \}$ denote a video of length $T$, where $I_t \in \mathbb{R}^{H \times W \times 3}$ is the frame at time step $t$. 
In the online setting, frames arrive sequentially. 
At time step $t$, the model only observes the prefix $\mathcal{V}_{\le t} = \{ I_1, \dots, I_t \}$.

\paragraph{Textual Query.}
Let $q$ denote a referring expression that specifies the target object. 
The query remains fixed throughout the video stream.

\paragraph{Online Segmentation Objective.}
At each time step $t$, the model predicts a binary mask $M_t \in \{0,1\}^{H \times W}$ corresponding to the object referred to by $q$ in frame $I_t$. 
Formally, an online model $f_\theta$ with parameters $\theta$ satisfies
\begin{equation}
M_t = f_\theta(\mathcal{V}_{\le t}, q),
\end{equation}
where the prediction at time $t$ must not depend on any future frame $I_{t'}$ with $t' > t$.

\paragraph{Causality Constraints.} 
In ORVOS, the causality constraint is a new yet important concept, which is to say that an object should be segmented iff. It satisfies the condition of the referring expression $q$. For instance, under the referring expression $q$ indicating `which watermelon has been cut', in ORVOS, the watermelon should be segmented only after it has been cut. The previous offline benchmarks aim to go through the entire video and then segment the watermelon before it is cut.

\paragraph{Evaluation.}
The predicted mask sequence $\{M_t\}_{t=1}^{T}$ is evaluated against ground-truth masks $\{M_t^{*}\}_{t=1}^{T}$ using standard video segmentation metrics under strictly causal inference.

\begin{figure*}[t]
\centering 
\includegraphics[width=0.9\columnwidth]{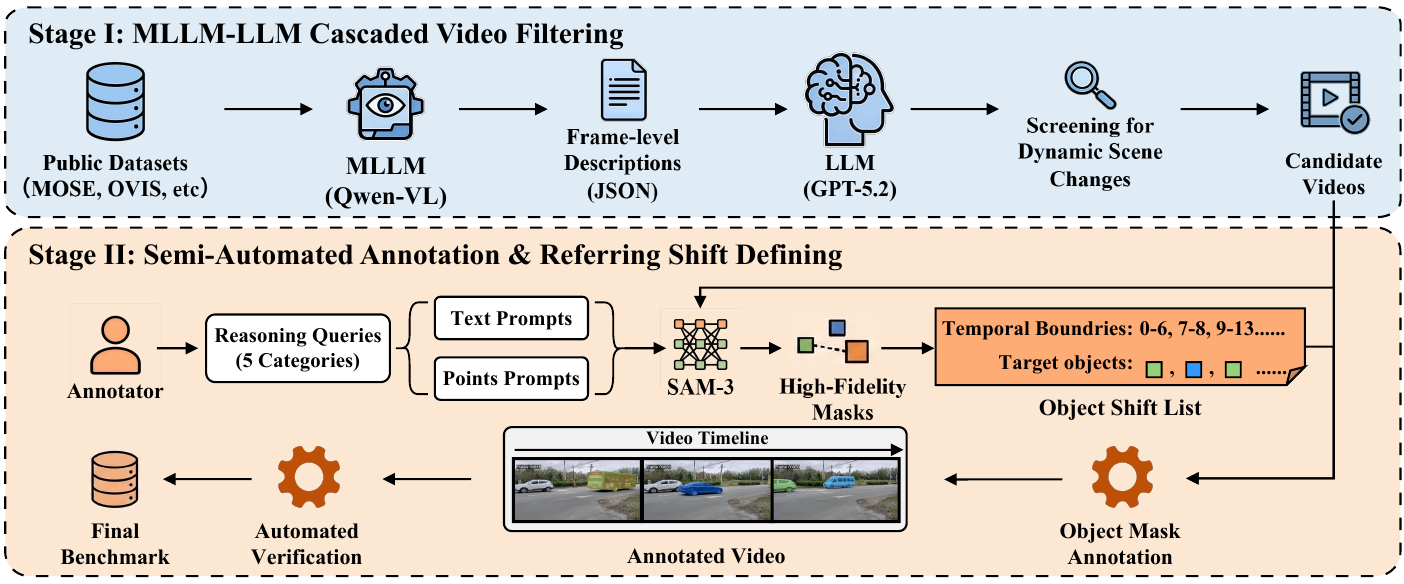}
  \caption{Overview of ORVOSB data construction and annotation pipeline.}
  \label{fig:orvsb_pipeline} \vspace{-15pt}
\end{figure*}

\subsection{Data Construction and Annotation}

Existing reasoning video object segmentation datasets typically assume a \textit{fixed target to be segmented} throughout the video, which is inadequate for evaluating online reasoning models where \textit{target identities dynamically shift} in tandem with unfolding events. 
To address this, we construct a new \textbf{evaluation} benchmark namely  ORVOSB (Online Reasoning Video Object Segmentation Benchmark), containing various \textit{referent shifts}. 

Given a referring expression, the referent shifts from one object to another in the video, which is not uncommon.
This way, we collect the raw video sequences from established datasets, including MOSE~\cite{ding2023mose}, OVIS~\cite{qi2022occluded}, YouTube-VIS 2022~\cite{yang2019video}, DAVIS17~\cite{pont20172017}, and MeViS~\cite{ding2023mevis}, by selecting videos that exhibit dynamic temporal transitions.
Based on the videos in the above datasets, we employ a two-stage MLLM-LLM cascaded automated screening pipeline to efficiently identify candidate sequences. 
First, a Multimodal Large Language Model (e.g., Qwen-VL~\cite{bai2023qwen}) processes the video frame sequences to generate detailed, frame-level textual descriptions formatted in JSON. Subsequently, these dense textual representations are fed into a Large Language Model (e.g., GPT-5.2) to identify sequences with potential dynamic scene changes. 
Human annotators then rapidly review these candidates. 
For each retained video and its corresponding query, we develop a human-in-the-loop annotation framework. 
We leverage the SAM-3~\cite{carion2025sam} model to generate initial high-fidelity segmentation masks based on human-provided prompts, with a small number of manual corrections. 
Crucially, to capture the \textit{referent shifts}, annotators carefully determine the start and end frames for each specific object identity as the video event unfolds. 
These temporal boundaries are recorded into a shift list, dynamically associating masks with the evolving query context. 
Finally, a verification script is employed to cross-check all annotated data, preventing missing frames of the mask annotation to ensure the temporal consistency.
The overall pipeline is shown in Fig.~\ref{fig:orvsb_pipeline}.

\subsection{Dataset Statistics and Analysis}
\begin{figure*}[h]
\vspace{-10pt}
\centering  
\includegraphics[width=1\columnwidth]{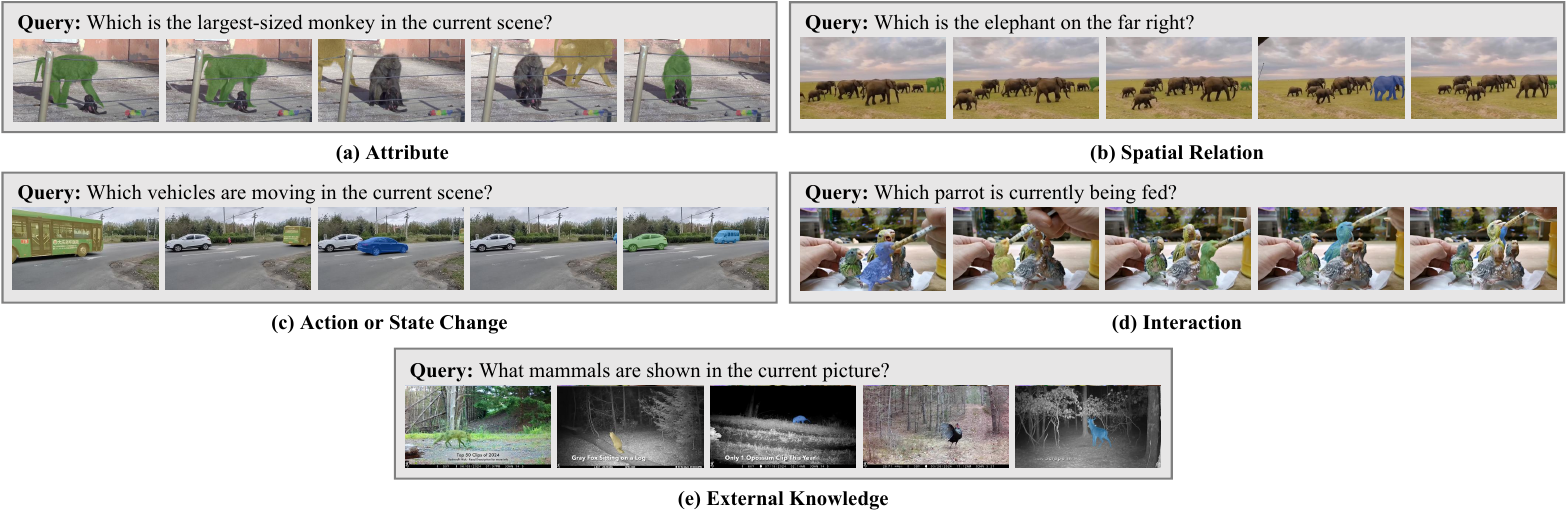}
  \caption{Illustrative examples of the five reasoning query types in ORVOSB.} 
  \label{fig:reasoning_categories} \vspace{-10pt}
\end{figure*}

In total, the final constructed benchmark comprises 512 evaluation samples, which come from 210 diverse video sequences, yielding a total of 12,907 annotated frames.
The video lengths vary significantly to reflect real-world streaming scenarios, ranging from a minimum of 12 frames to a maximum of 342 frames, with an average length of 61.46 frames per sequence. 
In this benchmark, we consider five types of reasoning query: attribute, spatial relation, action or state change, interaction, and external knowledge, which are shown in Fig~\ref{fig:reasoning_categories}.

Unlike static offline referring segmentation benchmarks, our dataset uniquely emphasizes dynamic referring target transitions. 
On average, each query contains 3.66 referent shifts, meaning the target identity changes multiple times within a single streaming sequence. Furthermore, 55.86\% of the queries involve scenarios of discontinuous target grounding, where an object temporarily loses its target status due to unfolding events but later regains it, as shown in Fig~\ref{fig:reasoning_categories}(a). This requires the model to maintain long-term representations of inactive objects, posing a severe challenge to the memory mechanisms of online reasoning models.
This distribution ensures that our benchmark comprehensively evaluates a model's ability to perform causality-grounded, event-aware object segmentation under streaming video input.

\section{Method}
\label{sec:method}
\subsection{Overview}

\begin{figure*}[h]
\vspace{-15pt}
\centering 
\includegraphics[width=0.9\columnwidth]{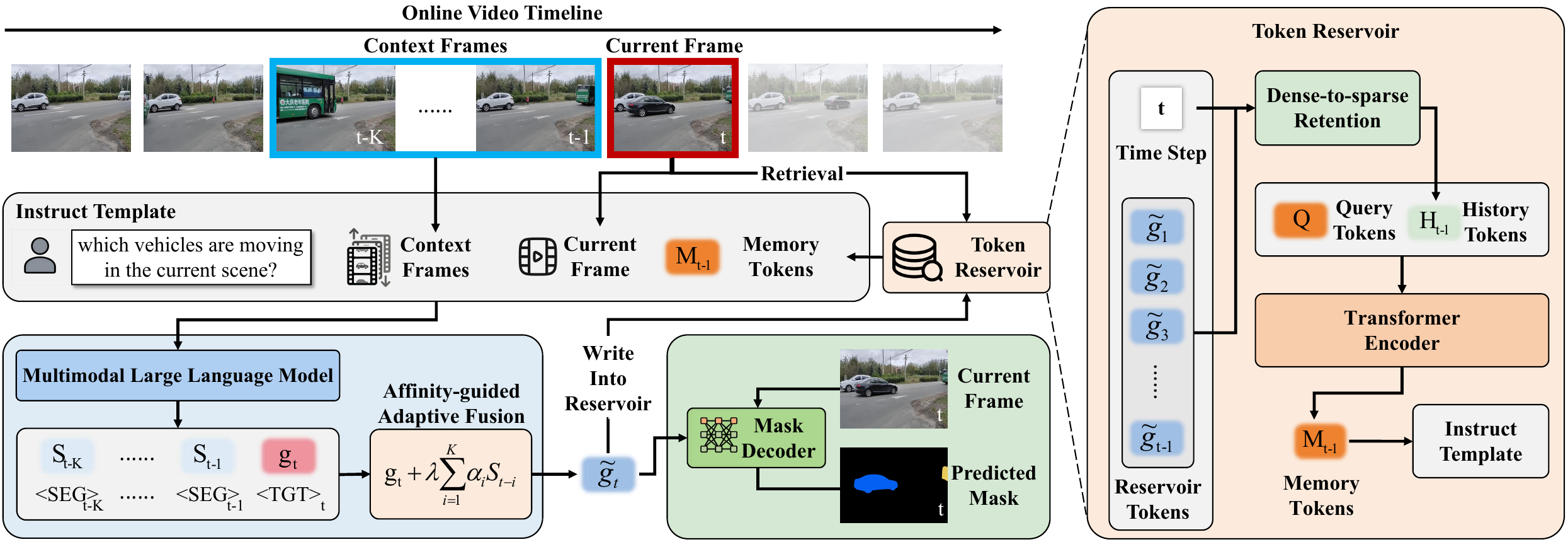}
    \caption{
    Overview of the proposed online reasoning video object segmentation framework. 
    At each time step, the model processes context frames and the current frame together with a dynamic token reservoir. 
    The multimodal large language model produces segmentation tokens, which are adaptively fused to form a history-aware prompt for mask prediction. 
    The fused token features are written back to the token reservoir, enabling structured long-term reasoning under strict causal constraints.
    }
    \label{overview} \vspace{-15pt}
\end{figure*}

Given a streaming video $\mathcal{V} = \{I_1, \dots, I_T\}$ and a textual query $q$, 
our goal is to perform causal reasoning video object segmentation under strict online constraints. 
To this end, we propose a memory-enhanced multimodal framework that integrates a multimodal large language model (MLLM), a dynamic token reservoir, and a segmentation head into a unified architecture, as illustrated in Fig.~\ref{overview}.

At each time step $t$, the model processes the current frame $I_t$, a short prefix of recent context frames $I_{t-K:t-1}$ (in terms of a K-frame sliding window), a dynamic token reservoir $\mathcal{M}_{t-1}$, and the query $q$. 
The MLLM generates frame-level segmentation tokens for context frames $I_{t-K:t-1}$ and a target-aware token for the current frame $t$. 
These tokens are adaptively fused to construct a history-aware query representation, 
allowing the segmentation prompt to evolve over time as more evidence is accumulated (Section~\ref{sec:prompt}). 
The resulting representation is fed into the segmentation head to predict the mask $M_t$, 
and the fused token-level features are subsequently stored in the `token reservoir' for future steps.
Here, to ensure stable long-term reasoning under streaming settings, the token reservoir adopts a temporally-aware strategy that retains denser representations for recent observations while progressively sparsifying distant history (Section~\ref{sec:memory}). 
This design enables the model to leverage both short-term temporal context and structured long-term memory while strictly preserving causality during online inference.

\subsection{Continually-Updating Segmentation Prompt}
\label{sec:prompt}

\paragraph{MLLM encoding and token extraction.}
At time step $t$, we construct the multimodal input sequence according to a predefined prompt template:
\begin{equation}
\mathbf{X}_t =
[\mathbf{T}_\mathrm{inst};\ \mathbf{T}_q;\ \mathbf{T}_\mathrm{vis};\ \mathbf{T}_\mathrm{mem}],
\end{equation}
where $\mathbf{T}_\mathrm{inst}$ denotes fixed instruction tokens,
$\mathbf{T}_q$ is the tokenized textual query, $\mathbf{T}_\mathrm{vis}$ are visual tokens extracted from the  context frames $I_{t-K:t-1}$ and current frame $I_{t}$,
and $\mathbf{T}_\mathrm{mem} = \mathbf{M}_{t-1}$ are memory tokens from the token reservoir (Section~\ref{sec:memory}).
The instruction template further includes predefined placeholder tokens
\texttt{<TGT>} and \texttt{<SEG>} at fixed positions,
serving as semantic anchors for the current target frame and context frames.

The sequence is processed autoregressively by the MLLM,
yielding contextualized embeddings
\begin{equation}
\mathbf{E}_t = \mathrm{MLLM}(\mathbf{X}_t)
\in \mathbb{R}^{L_t \times d}.
\end{equation}
As shown in Fig.~\ref{overview}, we read the predefined placeholder positions from the $L_t$ contextualized embeddings and obtain:
the embedding at position \texttt{<TGT>} forms the target-aware feature
$\mathbf{g}_t \in \mathbb{R}^{d}$,
while embeddings at positions \texttt{<SEG>} form context segmentation features
$\{\mathbf{s}_{t-i}\}_{i=1}^{K} \in \mathbb{R}^{K \times d}$.
This design allows the MLLM to encode target and contextual information in a unified sequence,
while preserving explicit structural roles within the prompt.

\paragraph{Affinity-guided adaptive fusion.}
To adaptively incorporate historical evidence,
we compute the affinity between the current target token and historical segmentation tokens:
\begin{equation}
u_i = \cos(\mathbf{s}_{t-i}, \mathbf{g}_t),
\qquad
\alpha_i = \frac{\exp(u_i)}{\sum_{j=1}^{K}\exp(u_j)}.
\end{equation}
The fused prompt representation is obtained by injecting weighted contextual information into the target token:
\begin{equation}
\tilde{\mathbf{g}}_t
=
\mathbf{g}_t
+
\lambda \sum_{i=1}^{K}\alpha_i\,\mathbf{s}_{t-i}.
\label{eq:tokenfusion}
\end{equation}
where  $\lambda$ is a pre-set parameter. This affinity-guided fusion emphasizes semantically relevant historical cues
while suppressing irrelevant context,
leading to a history-aware target representation.

\subsection{Structured Temporal Token Reservoir}
\label{sec:memory}

In online video reasoning segmentation, the model should reason over as much temporal context as possible under strict causality while considering a bounded computational budget. Storing all historical features causes unbounded memory growth, while naive truncation may discard critical semantic evidence. Furthermore, recent frames typically provide precise spatial cues, whereas distant frames contribute complementary high-level semantics. Motivated by these observations, we design a structured temporal token reservoir that compactly summarizes past information with temporally varying importance.

At time $t-1$, after obtaining the fused prompt feature $\tilde{\mathbf{g}}_{t-1}$ in Section~\ref{sec:prompt}, we maintain a token reservoir storing semantic features from \textit{all} previous steps.
Let $\tilde{\mathbf{g}}_\tau \in \mathbb{R}^{d}$ denote the feature written at time step $\tau$ ($\tau \leq t-1$),
and define the token reservoir
\begin{equation}
\mathcal{B}_{t-1}=\{\tilde{\mathbf{g}}_1,\dots,\tilde{\mathbf{g}}_\tau,\dots,\tilde{\mathbf{g}}_{t-1}\}.
\end{equation}

\paragraph{Dense-to-sparse retention.}
With the increase of time, storing \textit{all} previous steps is high-cost. To balance recency and long-term coverage, when $|\mathcal{B}_{t-1}| > N_{\max}$, we retain a fixed-length memory by selecting a temporally non-uniform index set $\mathcal{I}_{t-1}\subseteq\{1,\dots,t-1\}$, which samples recent steps more densely and distant steps more sparsely. Concretely, we construct $\mathcal{I}_{t-1}$ via a nonlinear time-warping function and map uniformly spaced coordinates to discrete time indices (Algorithm~\ref{alg:non_uniform_sampling}). The resulting sampled memory sequence is
\begin{equation}
\mathbf{H}_{t-1} =
\left[\mathbf{h}_{i}\right]_{i\in\mathcal{I}_{t-1}}
\in \mathbb{R}^{N \times d},
\end{equation}
where $N = N_{\max}$ when $|\mathcal{B}_{t-1}| > N_{\max}$, and $N = |\mathcal{B}_{t-1}|$ otherwise.

\vspace{-15pt}
\begin{algorithm}[h] 
\caption{Dense-to-Sparse Temporal Sampling}
\label{alg:non_uniform_sampling}
\KwIn{Current step $n$, maximum memory length $N_{\max}$}
\KwOut{Sampled index set $\mathcal{I} \subseteq \{1,\dots,n\}$}
\If{$n \le N_{\max}$}{
    \Return $\{1,2,\dots,n\}$\;
}
Initialize $\mathcal{I} \leftarrow \emptyset$\;
\For{$k = 0$ \KwTo $N_{\max}-1$}{
    $u \leftarrow \frac{k}{N_{\max}-1}$ \tcp*{Uniform coordinate in $[0,1]$}
    $\phi(u) \leftarrow 1 - (1-u)^2$ \tcp*{Nonlinear time warping}
    $t_k \leftarrow \lfloor \phi(u)(n-1) \rfloor + 1$ \tcp*{Map to $\{1,\dots,n\}$}
    Add $t_k$ to $\mathcal{I}$\;
}
\Return $\mathcal{I}$\;
\end{algorithm} \vspace{-20pt}

\paragraph{Query-based aggregation.}
Besides history token $\mathbf{H}_{t-1}$, we further introduce $L$ learnable memory query tokens
$\mathbf{Q}\in\mathbb{R}^{L\times d}$.
We add sinusoidal positional encoding $\mathbf{P}$ to $\mathbf{H}_{t-1}$ and jointly aggregate queries and history by applying a lightweight Transformer encoder $\mathrm{Agg}(\cdot)$ as
\begin{equation}
\mathbf{M}_{t-1} =
\mathrm{Agg}\big([\mathbf{Q};\ \mathbf{H}_{t-1}+\mathbf{P}]\big)_{1:L}
\in \mathbb{R}^{L\times d}.
\end{equation}
The resulting \textbf{memory tokens} $\mathbf{M}_{t-1}$ serve as an external mnemonic input to the MLLM.

Note that, the initial memory (i.e., $t=1$) tokens are obtained from $\mathbf{Q}$ only.

\paragraph{Memory update.}
As shown in Fig.~\ref{overview}, for current time $t$, we write $\tilde{\mathbf{g}}_t$ into the token reservoir:
\begin{equation}
\mathcal{B}_t = \mathcal{B}_{t-1} \cup \{\tilde{\mathbf{g}}_t\}.
\end{equation}



\subsection{Prompt-Guided Mask Decoding}

The fused prompt $\tilde{\mathbf{g}}_t$ is projected into the mask decoder embedding space via a lightweight MLP.
Given visual features $\mathbf{F}_t$ extracted from the current frame $I_t$,
the segmentation head predicts the mask:
\begin{equation}
M_t = \mathcal{D}(\mathbf{F}_t, \mathbf{z}_t),
\end{equation}
where $\mathbf{z}_t$ denotes the projected prompt feature of $\tilde{\mathbf{g}}_t$.

By conditioning mask prediction on the evolving prompt representation,
the model integrates visual evidence and accumulated semantic memory
while preserving online causality.

\subsection{Training and Implementation}
\subsubsection{Training Objective.}
We train the proposed framework end-to-end under a streaming protocol.
Given a video $\mathcal{V}=\{I_1,\dots,I_T\}$ and a query $q$,
we unroll the model along time.
At each step $t$, the model takes the current temporal window $I_{t-K:t}$,
the aggregated memory tokens $\mathbf{M}_{t-1}$, and the query $q$,
and produces the predicted mask $M_t$ as well as the autoregressive text output ${Y}_t$.
We optimize a joint objective that combines token-level language modeling and mask supervision.
Specifically, the step-wise loss is
\begin{equation}
\mathcal{L}_t
=
\mathcal{L}_{\text{CE}}(Y_t,\hat{Y}_t)
+
\lambda_{\text{bce}}\,\mathcal{L}_{\text{BCE}}(M_t, \hat{M}_t)
+
\lambda_{\text{dice}}\,\mathcal{L}_{\text{DICE}}(M_t, \hat{M}_t),
\label{eq:mask_loss}
\end{equation}
where $\hat{Y}_t$ denotes the ground-truth token sequences,
and $\hat{M}_t$ is the ground-truth mask for frame $I_t$.
$\mathcal{L}_{\text{CE}}$ is the standard autoregressive cross-entropy loss~\cite{graves2013generating},
$\mathcal{L}_{\text{BCE}}$ is the pixel-wise binary cross-entropy~\cite{wang2021exploring},
and $\mathcal{L}_{\text{DICE}}$ is the soft Dice loss~\cite{milletari2016v} that encourages mask overlap.

\subsubsection{Implementation Details.}
\label{sec:impl}
We build on Chat-UniVi-7B-v1.5~\cite{jin2024chat} as the MLLM and adapt it via rank-8 LoRA~\cite{hu2022lora}. We freeze the Chat-UniVi backbone and optimize the LoRA adapters and task-specific modules (SAM-2 mask decoder, prompt-projection MLP, and token reservoir).
We use AdamW~\cite{loshchilov2017decoupled} with a learning rate of $3\times10^{-4}$.
We set $\lambda{=}0.1$ in Eq.~\eqref{eq:tokenfusion} and $\lambda_{\text{bce}}{=}2$, $\lambda_{\text{dice}}{=}0.5$ in Eq.~\eqref{eq:mask_loss}.
Training uses DeepSpeed~\cite{rasley2020deepspeed} for 9{,}000 iterations on 8 RTX 4090 GPUs, with batch size 1 per GPU and 16-step gradient accumulation (effective batch size 128).
Unless specified, the context window is $K{=}4$ frames.
The structured temporal token reservoir performs query-based aggregation with $L{=}32$ learnable query tokens via a 2-layer, 8-head Transformer encoder and retrieves up to 32 historical tokens.

\section{Experiments}
\label{sec:exp}

\subsection{Setup}

\subsubsection{Datasets.}
We train our model on a mixed corpus of image and video data, combining segmentation supervision with multimodal instruction and video question answering data, including LLaVA-Instruct150k~\cite{liu2023visual} and the video QA datasets from Video-ChatGPT~\cite{maaz2024video}. For segmentation, we use standard image segmentation and referring segmentation datasets (e.g., ADE20K~\cite{zhou2017scene}, COCO-Stuff~\cite{caesar2018coco}, PACO-LVIS~\cite{ramanathan2023paco}, PASCAL-Part~\cite{chen2014detect}, refCOCO/refCOCO+/refCOCOg~\cite{kazemzadeh2014referitgame,mao2016generation}, and ReasonSeg~\cite{lai2024lisa}), together with video segmentation benchmarks (e.g., Ref-Youtube-VOS~\cite{seo2020urvos}, Ref-DAVIS17~\cite{pont20172017}, MeViS~\cite{ding2023mevis}, and ReVOS~\cite{yan2024visa}). We evaluate our method on ReVOS and our proposed ORVOSB benchmark.

\subsubsection{Evaluation Metrics.}
We evaluate video object segmentation using region similarity $\mathcal{J}$ (mean IoU) and contour accuracy $\mathcal{F}$ (boundary F-measure), and report their average $\mathcal{J}\&\mathcal{F}$ as the primary metric following standard practice in reasoning video object segmentation.

\subsection{Results on Online Benchmark}
We evaluate on ORVOSB and compare with the video-based methods VISA, VRS-HQ, GLUS, and VideoLISA, as well as the image-based baselines LISA and READ, as shown in Table~\ref{tab:category_analysis}.
VISA and VRS-HQ achieve only 33.3\% and 43.4\% $\mathcal{J}\&\mathcal{F}$, as their segment-then-track paradigm fixes the target on key frames and propagates it forward, making it difficult to handle referent shifts where the queried identity changes over time.
In contrast, LISA and READ achieve 46.6\% and 51.5\% $\mathcal{J}\&\mathcal{F}$ by constructing frame-specific segmentation prompts, allowing the grounding representation to adapt to the evolving temporal context.
VideoLISA reaches 50.3\% $\mathcal{J}\&\mathcal{F}$; however, it does not explicitly incorporate a causality-preserving prompt update mechanism for sequential target re-grounding, which is crucial in the online setting.
Our method achieves the best overall performance at 61.3\% $\mathcal{J}\&\mathcal{F}$, surpassing READ by +9.8\% and VideoLISA by +11.0\% with consistent gains across all reasoning categories, validating the benefits of continually-updating segmentation prompt and structured temporal token reservoir for progressive referent grounding under strict causality.

\begin{table*}[h]
\vspace{-15pt}
  \caption{Performance comparison with previous methods on the ORVOSB. 
Type indicates whether a method is image-based (Img) or video-based (Vid).}
  \label{tab:category_analysis}
  \centering
  \resizebox{0.85\textwidth}{!}{
  \begin{tabular}{l | c | c |c |c |c |c |c c c c}
    \toprule
    \multirow{2}{*}{Methods} &
    \multirow{2}{*}{Type} 
    & Attr. & Spatial & Action & Interact. & Ext. Know.
    & \multicolumn{3}{c}{Overall} \\
   
    \cmidrule(lr){3-10}
    & & $\mathcal{J}\&\mathcal{F}$ & $\mathcal{J}\&\mathcal{F}$ & $\mathcal{J}\&\mathcal{F}$ & $\mathcal{J}\&\mathcal{F}$ & $\mathcal{J}\&\mathcal{F}$ & $\mathcal{J}$ & $\mathcal{F}$ & $\mathcal{J}\&\mathcal{F}$ \\
    \midrule
    VISA~\cite{yan2024visa}       & Vid & 41.6\% & 31.0\% & 40.3\% & 27.5\% & 39.1\% & 30.8\% & 35.7\% & 33.3\% \\
    VRS-HQ~\cite{gong2025devil}     & Vid & 44.4\% & 44.3\% & 41.8\% & 35.8\% & 43.7\% & 40.4\% & 46.4\% & 43.4\% \\
    GLUS~\cite{lin2025glus}       & Vid &43.0\% 	&45.8\% 	&47.0\% 	&32.9\% 	&51.1\% 		&42.7\% 	&47.7\%   &45.2\% 
 \\
    LISA~\cite{lai2024lisa}       & Img & 53.6\% & 46.4\% & 47.7\% & 35.1\% & 49.5\% & 44.7\% & 48.4\% & 46.6\% \\
    VideoLISA~\cite{munasinghe2025videoglamm}  & Vid & 53.2\% & 52.1\% & 46.1\% & 34.0\% & 55.0\% & 47.5\% & 53.1\% & 50.3\% \\
    READ~\cite{qian2025reasoning}       & Img & 56.2\% & 52.4\% & 50.7\% & 38.6\% & 48.9\% & 49.8\% & 53.2\% & 51.5\% \\
    \midrule
    Ours       & Vid & ~\textbf{63.8\%} & ~\textbf{64.9\%} & ~\textbf{52.4\%} & ~\textbf{40.0\%} & ~\textbf{62.0\%} & ~\textbf{58.3\%} & ~\textbf{64.2\%} & ~\textbf{61.3\%} \\
    \bottomrule
  \end{tabular}}
\vspace{-15pt}
\end{table*}

\subsection{Ablation Study}
We conduct ablation studies on ReVOS and ORVOSB to evaluate the contribution of each module in our framework, as shown in Table~\ref{tab:ablation_cusp_tr}. The baseline processes only the current frame (and the same context window) into the MLLM, but uses only the current-frame target token as the segmentation prompt.
Introducing the continually-updating segmentation prompt (CUSP) yields consistent gains, improving $\mathcal{J}\&\mathcal{F}$ from 52.0\% to 53.0\% on ReVOS and from 52.4\% to 55.1\% on ORVOSB.
Removing affinity-guided adaptive fusion from CUSP slightly reduces the gain, indicating that adaptive fusion better leverages contextual cues.
Adding a token reservoir (TR) with uniform retention further improves $\mathcal{J}\&\mathcal{F}$ to 54.1\% on ReVOS and 60.9\% on ORVOSB, demonstrating the benefit of explicit historical evidence accumulation.
Replacing uniform retention with dense-to-sparse retention achieves the best $\mathcal{J}\&\mathcal{F}$ of 54.3\% on ReVOS and 61.3\% on ORVOSB, balancing recent detail preservation and long-term coverage under a bounded memory budget.
Overall, temporal prompt evolution and structured memory are complementary, and their combination performs best.

\begin{table*}[h]
\vspace{-10pt}
  \caption{Ablation study on ReVOS and ORVOSB. CUSP refers to the continually-updating segmentation prompt, AGAF to affinity-guided adaptive fusion, TR to token reservoir, US to uniform sampling retention, and D2S to dense-to-sparse retention.}
  \label{tab:ablation_cusp_tr}
  \centering
  \resizebox{0.65\textwidth}{!}{
  \begin{tabular}{l | ccc | ccc}
    \toprule
    \multirow{2}{*}{Methods} 
    & \multicolumn{3}{c|}{ReVOS} 
    & \multicolumn{3}{c}{ORVOSB} \\
    \cmidrule(lr){2-4}\cmidrule(lr){5-7}
    & $\mathcal{J}$ & $\mathcal{F}$ & $\mathcal{J}\&\mathcal{F}$ & $\mathcal{J}$ & $\mathcal{F}$ & $\mathcal{J}\&\mathcal{F}$ \\
    \midrule
    Baseline                              & 49.5\% & 54.5\% & 52.0\% & 49.2\% & 55.6\% & 52.4\% \\
    \midrule
    + CUSP (w/o AGAF)             & 50.5\% & 55.4\% & 53.0\% & 51.8\% & 58.3\% & 55.1\% \\
    + CUSP                  & 51.3\% & 56.1\% & 53.7\% & 55.0\% & 61.3\% & 58.1\% \\
    + CUSP  + TR (US)  & 51.8\% & 56.4\% & 54.1\% & 58.0\% & 63.9\% & 60.9\% \\
    + CUSP  + TR (D2S)      & \textbf{52.0\%} & \textbf{56.6\%} & \textbf{54.3\%} 
                                          & \textbf{58.3\%} & \textbf{64.2\%} & \textbf{61.3\%} \\
    \bottomrule
  \end{tabular}}
  \vspace{-20pt}
\end{table*}

\subsection{Comparison with Offline Methods on General Benchmark}
We evaluate on ReVOS and compare with prior methods, as shown in Table~\ref{tab:revos_full}. Our method achieves 54.3\% $\mathcal{J}\&\mathcal{F}$ overall, which is comparable to GLUS at 54.8\% $\mathcal{J}\&\mathcal{F}$ and substantially improves over earlier baselines such as LISA and TrackGPT. VRS-HQ attains the best overall performance at 59.1\% $\mathcal{J}\&\mathcal{F}$. Notably, both GLUS and VRS-HQ incorporate global video understanding, which is particularly effective on ReVOS where the referent remains fixed throughout the sequence and long-range context can be consolidated into a globally consistent representation. In contrast, we evaluate our model under a causal online constraint, producing predictions sequentially without access to future frames. This setting reduces the ability to leverage full-video cues, yet our method remains highly competitive, indicating that continually-updating segmentation prompt and structured temporal token reservoir can accumulate sufficient evidence over time and transfer robustly to conventional offline benchmarks even under stricter inference constraints.

\begin{table*}[h]
\vspace{-5pt}
  \caption{Performance comparison with previous methods on the ReVOS dataset.}
  \label{tab:revos_full}
  \centering
  \resizebox{0.7\textwidth}{!}{
  \begin{tabular}{l | ccc | ccc |  ccc}
    \toprule
    \multirow{2}{*}{Methods} 
    & \multicolumn{3}{c|}{Referring} 
    & \multicolumn{3}{c|}{Reasoning} 
    & \multicolumn{3}{c}{Overall} \\
    \cmidrule(lr){2-4} \cmidrule(lr){5-7} \cmidrule(lr){8-10}
    & $\mathcal{J}$ & $\mathcal{F}$ & $\mathcal{J}\&\mathcal{F}$ & $\mathcal{J}$ & $\mathcal{F}$ & $\mathcal{J}\&\mathcal{F}$ & $\mathcal{J}$ & $\mathcal{F}$ & $\mathcal{J}\&\mathcal{F}$ \\
    \midrule
    LISA~\cite{lai2024lisa}      & 44.3\% & 47.1\% & 45.7\% & 33.8\% & 38.4\% & 36.1\% & 39.1\% & 42.7\% & 40.9\%\\
    TrackGPT~\cite{zhu2023tracking}  & 46.7\% & 49.7\% & 48.2\% & 36.8\% & 41.2\% & 39.0\% & 41.8\% & 45.5\% & 43.6\% \\
    VISA~\cite{yan2024visa}      & 49.2\% & 52.6\% & 50.9\% & 40.6\% & 45.4\% & 43.0\% & 44.9\% & 49.0\% & 46.9\%\\
    GLUS~\cite{lin2025glus}      & 56.3\% & 60.7\% & 58.3\% & 48.8\% & 53.9\% & 51.4\% & 52.6\% & 57.3\% & 54.8\%\\
    VRS-HQ~\cite{gong2025devil}    &  {59.8\%}  &  {64.5\%}  &  {62.1\%}  
              &  {53.5\%}  &  {58.7\%}  &  {56.1\%}  
              &  {56.6\%}  &  {61.6\%}  &  {59.1\%}   \\
    \midrule
    Ours      & 55.7\% & 59.9\% & 57.8\% & 48.4\% & 53.3\% & 50.8\% & 52.0\% & 56.6\% & 54.3\% \\
    \bottomrule
  \end{tabular}}
\end{table*}

\subsection{Qualitative Results}
\begin{figure}[h]
\vspace{-10pt}
  \centering
  \includegraphics[width=1\linewidth]{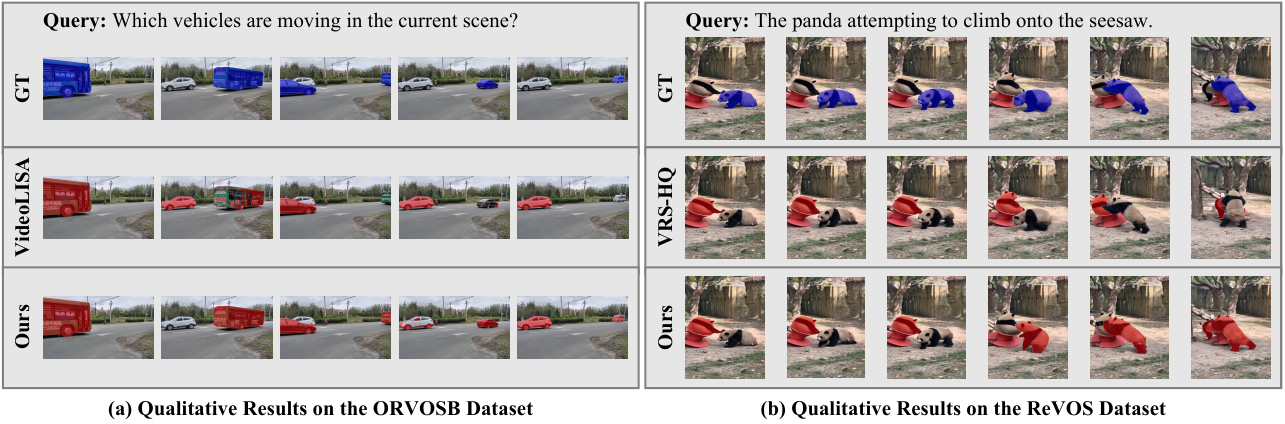}
  \caption{Qualitative comparison of segmentation results on ORVOSB and ReVOS.}
  \label{fig:qualitative_results}
  \vspace{-20pt}
\end{figure}

Fig.~\ref{fig:qualitative_results} presents qualitative comparisons on ORVOSB and ReVOS against strong baselines.
On ORVOSB, we compare with VideoLISA under queries involving dynamically evolving referents. As shown in Fig.~\ref{fig:qualitative_results}(a), VideoLISA tends to produce inconsistent masks when the queried identity changes over time, since its segmentation representation is largely shared across frames and lacks explicit temporal adaptation. In contrast, our method maintains stable and accurate masks as the referent evolves. This advantage stems from the continually-updating segmentation prompt, where the fused prompt token is updated at each time step, enabling the model to inject newly observed temporal evidence into the segmentation representation and adapt to dynamic changes.
On ReVOS, we compare with VRS-HQ as shown in Fig.~\ref{fig:qualitative_results}(b). VRS-HQ relies on a segment-then-track paradigm that determines the target on key frames and propagates it forward. Once an early prediction is suboptimal, subsequent masks may drift without correction. In contrast, our framework progressively updates prompt embedding and the token reservoir accumulates disambiguation cues over time. This design allows the model to refine its target representation sequentially and recover from ambiguous early frames, resulting in more temporally consistent segmentation.


\section{Conclusion}
In this paper, we introduce Online Reasoning Video Object Segmentation (ORVOS), a causally grounded setting that requires frame-by-frame prediction without access to future observations. In contrast to conventional offline RVOS, ORVOS prohibits retrospective disambiguation and instead demands incremental interpretation, particularly under referent shifts.
To support this formulation, we present ORVOSB, a benchmark with frame-level causal annotations and referent-shift labels for systematic evaluation of online reasoning and dynamic target adaptation. We further establish a strong baseline that maintains history-aware segmentation representations via continual prompt updates and structured temporal evidence aggregation under bounded computation.
Experiments show that existing RVOS methods degrade substantially in the online regime, underscoring the importance of causality-preserving temporal modeling for future research.

\clearpage  


%
%
\bibliographystyle{splncs04}
\bibliography{main}
\end{document}